*Review*

# A Scoping Review of Energy-Efficient Driving Behaviors and Applied State-of-the-Art AI Methods

Zhipeng Ma *, Bo Nørregaard Jørgensen * and Zheng Ma *

SDU Center for Energy Informatics, The Maersk Mc-Kinney Moller Institute, Faculty of Engineering, University of Southern Denmark, DK-5230 Odense, Denmark
* Correspondence: zhma@mmmi.sdu.dk (Z.M.); bnj@mmmi.sdu.dk (B.N.J.); zma@mmmi.sdu.dk (Z.M.)

**Abstract:** The transportation sector remains a major contributor to greenhouse gas emissions. The understanding of energy-efficient driving behaviors and utilization of energy-efficient driving strategies are essential to reduce vehicles' fuel consumption. However, there is no comprehensive investigation into energy-efficient driving behaviors and strategies. Furthermore, many state-of-the-art AI models have been applied for the analysis of eco-friendly driving styles, but no overview is available. To fill the gap, this paper conducts a thorough literature review on ecological driving behaviors and styles, and analyzes the driving factors influencing energy consumption and state-of-the-art methodologies. With a thorough scoping review process, thirty-seven articles with full text were assessed, and the methodological and related data are compared. The results show that the factors that impact driving behaviors can be summarized into eleven features including speed, acceleration, deceleration, pedal, steering, gear, engine, distance, weather, traffic signal, and road parameters. This paper finds that supervised/unsupervised learning algorithms and reinforcement learning frameworks have been popularly used to model the vehicle's energy consumption with multi-dimensional data. Furthermore, the literature shows that the driving data are collected from either simulators or real-world experiments, and the real-world data are mainly stored and transmitted by meters, controller area networks, onboard data services, smartphones, and additional sensors installed in the vehicle. Based on driving behavior factors, driver characteristics, and safety rules, this paper recommends nine energy-efficient driving styles including four guidelines for the drivers' selection and adjustment of the vehicle parameters, three recommendations for the energy-efficient driving styles in different driving scenarios, and two subjective suggestions for different types of drivers and employers.

**Keywords:** driving behavior; driving pattern; energy efficiency; artificial intelligence; literature review

## 1. Introduction

The $CO_2$ emissions in the transportation sector form a significant component of the manmade greenhouse gas (GHG) which results in global warming through the greenhouse effect. While this sector gradually decarbonizes, it still contributes to almost 30% of GHG emissions [1,2], 65% of which are caused by road transport [3]. Therefore, the strategies to improve fuel economy need to be studied to reduce $CO_2$ emissions.

Energy-efficient driving technology is an important factor in reducing vehicle fuel consumption. It refers to the decisions that a driver can make to improve the efficiency of the engine [4]. These decisions could be multidimensional, involving vehicle selection [5], route planning [6], and driving behavior recommendation [7]. Some industrial stakeholders, like the public transportation and logistics companies, already have enough vehicles, and the route should be planned based on not only the fuel economy, but also the work requirements and efficiency. The vehicle and route conditions can only be controlled by their manufacturers or constructors, and the drivers can only optimize their driving behaviors and styles to enhance energy efficiency. Therefore, driving behavior analysis is an important research domain to recommend the least fuel-consumption driving style in such industry sectors.

The relationship between fuel consumption and driving behavior has been a popular research topic recently with the development of artificial intelligence (AI) and machine learning (ML). The main factors representing and influencing driving behavior include velocity, acceleration, gear, road parameters, weather, etc., and the data can be collected through sensor networks and CANbus (controller area network) [8]. With the ever-increasing volume of generated data, traditional models like linear regression cannot produce accurate estimation results in real-world applications [9]. In recent years, many state-of-the-art machine learning (ML) (e.g., random forest [10] and neural networks [11]) and reinforcement learning (RL) [12] models have been developed to perform research on energy consumption. Random forest is utilized in [10] to classify the road types and more precise environment perception helps to enhance vehicle energy efficiency. The neural networks model is applied in [11] to estimate the fuel consumption of three vehicles on distinct road conditions. The study in [12] utilizes an RL model to generate energy-saving driving behaviors and integrates vision-perceptive methods to achieve higher energy efficiency. Meanwhile, some studies apply classification [13] or clustering [14] algorithms to find out the best practices in driving styles for fuel saving.

Although some publications have introduced multiple AI models in the energy-efficient driving aspect, the comprehensive recommendations on driving behavior are not sufficient in the studies. The studies in [4,9] have made recommendations based on vehicle planning algorithms and vehicle maintenance, without considering the driving behaviors. Moreover, the recommendation for energy-efficient driving is unclear in many review papers. The study in [15] indicates that some energy-efficient driving rules do not fit the ingrained driving habits, which cannot make drivers understand the nature of energy-efficient driving properly. Hence, it is necessary to collect and analyze the articles concerning driving behavior data processing and fuel consumption optimization, as well as make clear and user-friendly recommendations for energy-efficient driving [16]. The scoping review is capable of finding out the limitations and challenges of current research on this topic, investigating applied theories and methodologies in the field which supports the selection of promising methods, and guiding the direction of future studies [17].

Therefore, this paper conducts a scoping review for investigating the state-of-the-art AI models and their applications to analyze the driving behavior impacts on fuel consumption. Specifically, this paper investigates and recommends the ML/RL algorithms to estimate fuel consumption based on driving behavior data and select the most ecological driving style. Moreover, the main factors concerning driver behaviors are also analyzed and the recommendations for energy-efficient driving are concluded in this paper.

The remainder of the paper is organized as follows: Section 2 describes the methodology and the research process of a scoping review. The literature analysis results are presented in Section 3. Section 4 highlights the findings about the state-of-the-art AI models and the main factors related to driving behavior. The recommendations for energy-efficient driving are stated in this section as well. Section 5 presents the conclusion, the limitations, and the recommendation for future work.

## 2. Methodology

This paper conducts the scoping review research to investigate the state-of-the-art AI models to study the relationship between driving behavior and energy-efficient driving. Unlike other literature review approaches, such as systematic review and meta-analysis, which focus on specific questions or purposes narratively, the scoping review aims to investigate all of the available literature on a topic area. A scoping review aims to map the searched literature on a specific research topic and provide an opportunity to identify the key concepts and research gaps in this topic area [18,19].

To retrieve the relevant articles in the fields of AI and the applications in energy-efficient driving, four databases, including the ACM digital library, IEEE Xplore, Web of Science, and Scopus, were selected. The selection criteria for choosing databases involve reliability, content stability and reportability. Firstly, the articles in scientific research should be peer-reviewed, and databases like Web of Science and Scopus have rigorous standards for selecting journals and articles, ensuring higher quality and peer-reviewed content. However, Google Scholar includes a wide array of materials, not all of which are peer-reviewed. Secondly, the database structure should be constant. Databases like Web of Science, Scopus, and others maintain more stable content, essential for the rigorous needs of scoping reviews. Google Scholar's content, algorithms, and database structure are constantly changing, which can impact the stability and consistency of search results over time. Lastly, the search results should be accurate and reproducible. Databases like Web of Science provide more control over searches, with structured and uniform data entry that ensures accurate and reproducible search results, crucial for systematic reviews. The search results on Google Scholar are not always reproducible due to its variable content and automated search criteria. This variability makes it difficult to report search results accurately in a systematic review. Therefore, this study selects the databases of ACM digital library, IEEE Xplore, Web of Science, and Scopus, and excludes Google Scholar.

The literature search was performed in June 2022, and there was no limitation on the publication years for the search. The following search string was designed and searched in the mentioned databases: ("data" OR "artificial intelligen*" OR "learn*") AND ("driv*" OR "vehicle*") AND ("eco" OR "energy effici*"). The first part of the search string outlines the methodology, emphasizing data-driven models and AI-related algorithms. The subsequent sections delineate the thematic scope, with a specific focus on energy-efficient driving behaviors, as indicated by the title of this article.

Because of the large amount of literature in this field, the keyword search was only applied to titles to reduce the irrelevant publications. A total of 682 articles were mined and imported to the reference management software Endnote 20. These 682 articles were not unique due to the overlaps of different databases. The search results from each database are shown in Table 1.

**Table 1.** Results from each database.

| Database | Result |
| --- | --- |
| ACM | 276 |
| IEEE | 72 |
| Web of Science | 175 |
| Scopus | 159 |
| **Total** | **682** |

After the automatic and manual duplication check, 410 articles remained for further analysis. Then, the title search was conducted again in Endnote with the keywords of "vehicle" or "driving", since the initial keyword "driv*" included driven or driver, which refers to "a factor which causes a particular phenomenon to happen or develop" and which is irrelevant to vehicle systems and driving behaviors. It resulted in 112 papers.

Next, the remaining 112 papers were investigated by evaluating whether the articles' abstracts matched the scope of this review. If any article was in the scope but poorly written, the article would be kept for full text searching and analysis. Because only the scope of an article needed to be identified during this step, the abstracts provided enough information regarding the relevance of the scope of a scientific publication [20]. This step resulted in 68 articles being removed and 44 articles remaining. Finally, the full text for 37 out of the 44 articles was found and downloaded in the reference management software for further detailed text analysis. Table 2 summarizes the AI models, and Table 3 concludes the influencing factors of driving behaviors in the 37 articles.

**Table 2.** Summary of AI models in searched 37 publications.

| Categories | AI Models | References |
|---|---|---|
| Regression | Linear regression | [7,13,21–25] |
| | ANN | [11,26] |
| | LSTM | [12,27] |
| | kNN | [14] |
| | Gaussian process | [28,29] |
| | Gradient decent | [30] |
| | Decision tree | [31,32] |
| | Ensemble learning | [32] |
| Classification | Gaussian mixture models | [7] |
| | Decision tree | [8,33] |
| | kNN | [10] |
| | SVM | [10] |
| | Random forest | [10,32] |
| | CNN | [12] |
| | Rule-based | [34] |
| | Hidden Markov model | [22] |
| Clustering | K-means | [14,22,35] |
| Reinforcement learning | Reinforcement learning | [12,36–47] |

**Table 3.** Summary of influencing factors in the 37 searched publications.

| Categories | Influencing Factors | References |
|---|---|---|
| Factors reflecting driving behaviors | Speed | [7–14,21–23,25–39,41–49] |
| | Acceleration | [7–10,12,21–28,30–34,36–38,41,42,49] |
| | Deceleration | [7–9,24,25,30,31,34,42] |
| | Pedal | [10,13,14,31] |
| | Steering | [23–25,39] |
| | Gear | [9,25,32,35,43,48] |
| | Engine | [7,9,13,21,33,43,48] |
| Factors impacting driving behaviors | Distance | [10,12,26,29,40] |
| | Weather | [8,9] |
| | Traffic signal | [12,28,36,37,40,46] |
| | Road | [9,25,32,38,40,44,49] |

The AI models in Table 2 represent all algorithms that can be trained on a set of collected data to make decisions on energy-efficient driving, including conventional ML algorithms such as linear regression, decision tree, and SVM, and state-of-the-art algorithms, such as ANN and LSTM. They were categorized into regression, classification, clustering, and RL algorithms, and the details are discussed in Section 3.5.

The influencing factors presented in Table 3 pertain specifically to driving behaviors, as the focus of this study is on energy-efficient driving behaviors. While vehicle-related factors undeniably play a significant role in energy consumption, it is important to note that they fall outside the scope of this study. The details are discussed in Section 3.4.

## 3. Results

The 37 selected articles were analyzed in detail to investigate the current research on the AI models for driving behavior analysis and energy-efficient driving. This section was divided into the following sub-sections: (1) overview of the energy-efficient driving research, (2) data and data sources for energy-efficient driving research, (3) factors impacting driving behaviors, and (4) AI models applied in energy-efficient driving research.

*3.1. Overview of Energy-Efficient Driving Research*

The energy-efficient driving technology represents the strategies to merge vehicle speed management approaches and GHG emission reduction techniques, aiming at minimizing fuel consumption [50]. Of course, the benefits of energy-efficient driving go beyond energy saving. For one thing, the driving cost to the individuals and the companies can be reduced. For another, when developing energy-efficient driving strategies, safety conditions are also included, so as to reduce accidents and traffic fatalities [51]. Hence, the goal of energy-efficient driving technology is to help the driver choose a $CO_2$-reduction driving strategy under some safety and law conditions [50].

Recent research reveals that fuel consumption can be reduced by approximately 15% under different optimization approaches and various road conditions. The GMM model is employed in [7] to learn different driving modes and the most energy-saving acceleration model is calculated to save up to 15.81% energy. The hybrid RL method proposed in [12] reduces energy consumption by 12.70%. The developed data-driven optimal energy consumption cost model and optimal battery current model [26] are, respectively, constructed via two neural networks and can improve fuel economy by up to 16.7%. There are different topics of energy-efficient driving deserving investigation, and distinct research aspects to optimize the energy-efficient driving strategies. Figure 1 summarizes the research on energy-efficient driving, and they are analyzed in two parts in the following discussion, including the popular research topics and the applied technical methodologies.

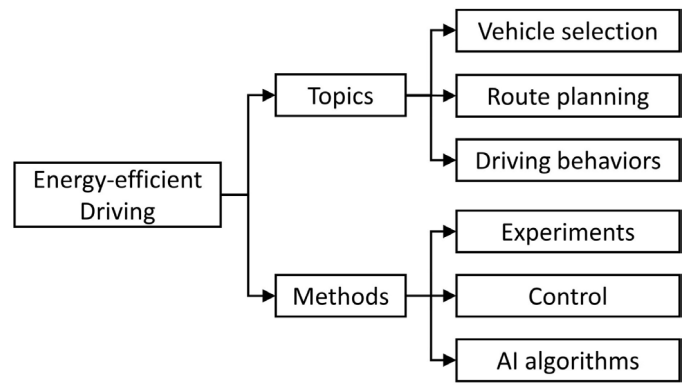

**Figure 1.** Summarization of energy-efficient driving research.

As Figure 1 demonstrates, the main topics in energy-efficient driving research include vehicle selection, route planning, and driving behaviors.

In the studies of vehicle selection, the relationship between vehicle parameters, including mass, the tire–road friction coefficient, power train parameters, etc., are analyzed, and the structure of the vehicle is optimized to save fuel consumption. For instance, neural networks are employed to map the efficiency of a planetary gearbox for an electric vehicle based on powertrain data generated from the efficiency experiments and design an energy-efficient powertrain in [52].

Vehicle route planning is another important study aspect. An energy-saving routing algorithm is capable of reducing the driving distance and frequency of acceleration/deceleration. A routing algorithm based on historical driving data to locate in energy-efficient routes is proposed by [53], where 51.5% of energy is saved in the case study. Lastly, ecological driving behavior plays an important role in fuel saving through the proper control of gas/brake/clutch pedals [54]. The driving-behavior prediction can guide the drivers to adjust the control style of the vehicle, avoiding inefficient driving. This paper narrows the research field of energy-efficient driving into the driving behavior, as the industrial stakeholders purchase the vehicles and design the routes based on many other demands besides energy efficiency.

The literature shows that the methods to investigate the relationship between energy consumption and driving behaviors can be divided into naturalistic experiments, control technologies and AI models. Although AI models are applied in both experiment and control studies as well, the majority of the reviewed articles analyze them from distinct perspectives. The experiment study focuses on designing experiments or physical simulators for conducting experiments and collecting data related to energy consumption. The control technologies in this article represent the control systems without applying any AI models.

Doing experiments in real-world environments or simulators is an intuitive approach to comparing energy consumption with different parameter settings. For instance, ecological driving behaviors and the relative prioritization of driver safety and energy-efficient driving in a driving simulator are studied in [55]. A networked game to collect large-scale human driving behavior data are developed in [56] to investigate ecological acceleration behavior.

Control technology is another popular topic in energy-efficient driving, in which the real-time control algorithm is developed to calculate an optimal control action, guiding the drivers or being utilized as a predictive cruise controller. For instance, a fuzzy logic control algorithm is proposed in [57] to compute the optimal velocity for a specific location, improving the fuel economy of vehicle fleets.

With the theoretical development of machine learning, many state-of-the-art AI models are applied in the energy and transportation sectors. Some prediction algorithms (e.g., random forest [10] and neural networks [26]) are applied to predict fuel consumption based on driving behavior data and select the best practices in energy-efficient driving. Meanwhile, some reinforcement learning frameworks [12,36] are proposed to analyze the ecological driving style in the simulation environment. The literature on experiments and control theory is excluded, since this paper focuses on AI algorithms in energy-efficient driving applications.

*3.2. Data Sources for Energy-Efficient Driving Research*

Before modeling driving behavior and studying its impact on fuel consumption, it is significant to collect the vehicle and driver data in real-world conditions. There are various data sources of driving behaviors, including simulation tools, meters, CANbus, OBDs (on-board diagnostics), and smartphones, which are shown in Table 4.

**Table 4.** Data sources for energy-efficient driving.

| Data Sources | | Counts | Ref. |
|---|---|---|---|
| Simulation data | | 14 | [21,27–29,36–42,44–46] |
| Embedded sensors | Meters | 5 | [14,22,26,30,47] |
| | CANbus | 6 | [13,22,23,35,43,49] |
| | OBDs | 5 | [7,11,12,31,33] |
| Smartphones | | 2 | [33,35] |
| Additional sensors | | 9 | [10–14,24,34,43,49] |

The easily implemented and efficient method is to collect data in the simulation environment so that multiple routes and climates can be set and multi-dimensional datasets are obtained easily. For instance, the driving performance data in a driving simulator is generated in [21], and the driving behavior data in a networked game are collected and analyzed in [56].

Although simulation data are easy to generate and abundant, they still differ from real-world data because some conditions might be simplified in the simulation systems. For example, the weather conditions and traffic congestion are difficult to reproduce in simulators. Hence, the data collected through naturalistic driving experiments is more convincing. The important variables connected to driving behavior are detected by some sensors and recorded by different devices. The sensors and devices embedded in the vehicles include meters, CANbus, and OBDs.

The basic data collection method is to read the odometer and log the fuel use, mileage, and velocity manually. This method is relatively simple and cheap, but human errors may occur in the data recording [4]. In addition, only a few observations and variables can be recorded manually, so the volume may not be big enough for big data analysis.

A more efficient data collection way is to utilize data loggers, which are plugged into CANbus and OBDs. CANbus is the controller area network bus, allowing microcontrollers to communicate with each other's applications without a host computer [58]. CANbus can provide detailed data concerning the running conditions of a vehicle, e.g., fuel consumption per second, real-time position, velocity, acceleration, and engine conditions. OBDs represent on-board diagnostic scanners and are usually connected to the engine control unit to provide real-time driving data [59]. The data loggers connected to the CANbus and OBDs are allowed to collect real-time data during normal driving. Moreover, they can be equipped to a large number of vehicles at a low cost. However, there is no standard for the design and manufacturing, and thus the collected variables vary in different products.

Furthermore, smartphones have been used for data collection recently. Many sensors and software (e.g., the GPS and the accelerometer) are equipped in smartphones, so they are capable of collecting most of the required data [60]. The weather and road conditions are available online and collected via smartphones as well. The combination of smartphones and dataloggers in CANbus or OBDs fulfills almost all the requirements of the input datasets. For instance, a dataset of 2250 driving tests with an Android smartphone and an OBD-Bluetooth adapter was generated in [54], and two ML models to analyze the dataset from CANbus and a smartphone, respectively, were developed in [50]. Some studies require customized data not covered in the mentioned sources and additional sensors (e.g., the camera [12], pedal force [14], and V2I communication devices [34]) should be installed for data collection.

### 3.3. Variables Reflecting Energy Efficiency

To assess energy efficiency, different variables are selected and computed in distinct research. Table 5 shows the four variables regarding energy efficiency and their units in different studies.

**Table 5.** Variables reflecting energy efficiency.

| Variables | Units |
|---|---|
| Fuel consumption | mL/s [13,49] <br> g [7] <br> L/km [8] <br> L/100 km [11,25,31,40] <br> gallon/mile [22] <br> mL [28,29,36] <br> kg [32] <br> gallon [44,45] |
| Electrical energy consumption | Wh/km [10,23] <br> J [23] <br> kwh/100 km [14,48] <br> kwh [30,47] <br> Wh [37] <br> kJ/s [42] |
| Fuel economy | km/L [24] <br> mile/gallon [39] |
| $CO_2$ emissions | g/km [21,48] <br> g/mile [22] <br> g [46] |

The variables are mainly divided into energy consumption and energy economy. Energy consumption signifies the amount of fuel or electricity a vehicle utilizes to cover a specific distance. There are many different measurements of energy consumption, as shown in Table 5. Specifically, gallon/mile and gallon are used in the USA [22,44,45]. Fuel economy denotes the distance a vehicle can travel on a defined quantity of fuel. In the USA, it is typically measured in miles/gallon [24], while in other countries, the standard unit is km/L [39]. Furthermore, certain studies employ direct measurements of $CO_2$ emissions to assess the environmental impact of driving practices, providing valuable insights for sustainable transportation and policy interventions aimed at reducing carbon emissions in the transportation sector.

### 3.4. Factors Impacting Energy-Efficient Driving Behaviors

As Table 3 demonstrates, various variables of driving behavior are collected in different studies. Eleven main factors can be concluded based on the literature, and each includes various sub-variables. These factors can be divided into two groups: factors reflecting driving behavior and factors affecting driving behavior. The factors reflecting driving behavior include the speed, acceleration, deceleration, pedal, steering, gear selection, and engine, which are controlled by the drivers. The factors affecting driving behavior represent the objective vehicle and environmental parameters, involving distance, weather-, traffic signal-, and road conditions. Their definitions are shown in Table 6.

**Table 6.** Definitions of influencing factors of energy-efficient driving.

| Categories | Influencing Factors | Definition |
|---|---|---|
| Factors reflecting driving behaviors | Speed | real-time linear velocity of the vehicle |
| | Acceleration | real-time acceleration of the vehicle |
| | Deceleration | real-time deceleration of the vehicle |
| | Pedal | (gas/brake/clutch) pedal force, pedal frequency, and pedal depth |
| | Steering | angle of the rotating steering wheel |
| | Gear | selection of gear ratio of a manual vehicle |
| | Engine | engine load and engine speed |
| Factors impacting driving behaviors | Distance | distance between vehicles, distance from vehicle to the traffic light infrastructure and distance from vehicle to the station |
| | Weather | temperature, visibility, rainfall, and wind speed |
| | Traffic signal | traffic signal status generated from the infrastructures |
| | Road | road geometry, road slope and radius of the curve of the road |

Distinct features are selected based on different demands and data collection methods in the literature. The counts of the occurrence of the factors are presented in Figure 2. Most of the studies analyzed speed and acceleration as they are the key targets of energy-efficient driving. The main purpose of energy-efficient driving is to develop a velocity and acceleration recommendation strategy for reducing energy consumption. Deceleration and acceleration are integrated in some studies [12,21,28], and the pedal indexes (e.g., the force, frequency and depth) are also the parameters indicating acceleration [14,40,47].

An optimized engine control strategy is another significant approach for fuel saving through the speed selection of the transmission box and the control of pedal force and depth. The gear shift and pedal control can affect the energy efficiency of the engine [43]. For instance, when the driver selects the small gear ratio in a high-speed work condition or presses the gas pedal quickly, the engine torque cannot match the velocity/acceleration perfectly and wastes a lot of energy. The engine parameters are easily collected from CANbus datasets, whereas the gear and pedal data collection requires additional sensors or manual recording in many studies. The data regarding gears and the position of the gas or brake pedals are recorded through additional sensors [10,43]. Except for direct measurement, these parameters can be computed based on the raw data recorded by OBD-II [31].

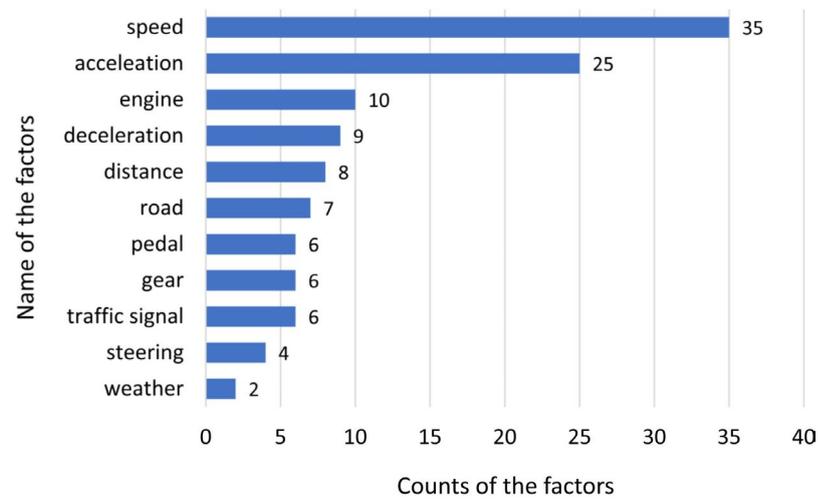

**Figure 2.** Counts of the influencing factors in the reviewed articles.

Distance is the most popular research point that impacts driving behavior, which contains multiple features, including the distances to the traffic lights or the station and those between the vehicles in different scenarios and rules [12,29,32]. For instance, drivers should control the brake pedal to wait for the green light or adjust the vehicle distance in different road conditions. Hence, the distance analysis is often accompanied by road conditions [12] and traffic signal status [25]. Moreover, weather conditions [8,9] influence driving styles significantly since the change of temperature and humidity results in the alteration of the coefficient of resistance, so that the speed limit and the relationship between acceleration and pedal depth are altered. However, weather detection can only be based on external sensors or the Internet, and such information is difficult to collect solely by the vehicle's own data collection equipment [9], so only two studies in the literature have integrated the weather data.

The combination of the mentioned features is also discussed in the literature. For instance, four types of influential variables including vehicle characteristics, driver characteristics, driving behavior, and weather conditions are summarized in [8], and a fuel-consumption classification model based on the decision tree was established to train the generated datasets. The driving behavior data from the CANbus dataset and questionnaires is collected in [23] with the velocity, acceleration, and steering wheel angle being generated from CANbus, while the questionnaire assembles the subjective driving characteristics (e.g., self-confidence, impatience, and rude driving behavior). The data from a driven vehicle, including the velocity, acceleration, pedal variables, engine parameters, road conditions, etc., are measured in [25], and a real-time fuel consumption estimation method is proposed for recommending the optimal speed in time. Furthermore, a deep-learning framework is developed in [11] to analyze the data from the OBD-II module and CANbus, where the input features include the velocity and engine parameters.

*3.5. AI Models Applied in Energy-Efficient Driving Research*

Two main strategies are applied in the energy-efficient driving analysis methodologies: (1) fuel consumption is estimated based on the input variables about driving behaviors, and the relationship between fuel consumption and driving behaviors is investigated through prediction algorithms; (2) RL frameworks are developed to build the multi-agent simulation environment to learn the optimal energy-efficient driving style. In our surveyed literature of 36 articles, 23 articles (64%) developed prediction models, whereas the rest (36%) proposed RL frameworks.

### 3.5.1. Prediction Models for Energy-Efficient Driving

The prediction models include regression and classification tasks. In regression tasks, the values of the target features (e.g., fuel consumption) are predicted by training a combination function based on the input variables [21,28]. In classification tasks, the driving styles are classified into several groups (e.g., energy-efficient and inefficient styles) [33,34]. Specifically, the unsupervised classification methods (the clustering methods) group the data based on their similarities [35]. As shown in Figure 3, 58% of the developed models in the literature are regression frameworks because driving is a continuous process, and regression results can provide more clear and accurate recommendations for energy-efficient driving than classification models [21].

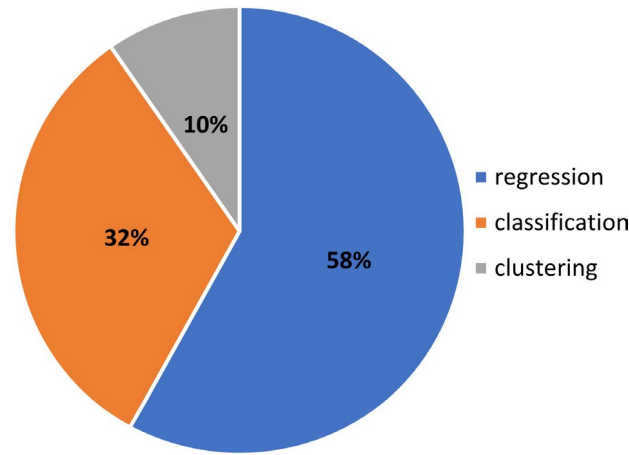

**Figure 3.** Distribution of applied AI models.

The linear regression family is the most popular model in the cases because it is a transparent and easy-implemented approach [2,7,13,18,21,23–25,27,40]. The contribution of each input variable can be recognized from the weights, allowing good mathematical interpretation [61]. For instance, bivariate regression methods are used in [13] to predict the relationships between each influencing factor and fuel consumption. The energy-efficient driving suggestions are made based on a series of regression lines. For example, the contributions of different energy-efficient driving rules to $CO_2$ emission reduction are measured through linear regression analysis in various scenarios in [21]. The result shows that fuel consumption is reduced by 12.38% in the guided-video scenario, and the proper use of the accelerator pedal is the most important factor affecting energy-efficient driving. Furthermore, the driving behaviors in both smooth and congested traffic scenarios are analyzed in [23]. The linear regression results demonstrate that proactive and rude driving wastes a lot of energy in smooth conditions, whereas the constant speed share and average acceleration are more significant in congested conditions.

However, the linear regression models have some drawbacks in their application due to their inner mechanisms. On one hand, the models do not always achieve high estimation accuracy because some relationships between influencing variables and energy consumption are not linear. For instance, the evolution of energy consumption is not linear-related to the speed since driving at the velocity of 50–80 km/h consumes less fuel than that at lower and higher speeds [9]. On the other hand, one condition of linear models is that there is no linear correlation between the input variables, otherwise the weights cannot demonstrate their influence on fuel consumption correctly, but some of the driving behavior features are strongly correlated (e.g., gas pedal and acceleration) [62].

Other regression algorithms that can process both linear and non-linear functional forms are used in other cases to avoid such limitations. For instance, a Gaussian process regression model is proposed in [29] to predict the signalized intersection crossing time for optimizing the speed trajectory. The proposed energy-efficient driving strategy saves energy by approximately 4% in the simulation environment. In another study [11], the fuel consumption was approximated through the Elman neural network, and the prediction accuracy was over 96%. The contributions of each variable can be analyzed through the feature importance detection algorithms. Furthermore, for example, a framework based on Shapley additive explanations (SHAP) was developed in [37] to indicate the significant variables affecting fuel consumption and design energy-efficient driving speed and acceleration strategies based on SHAP results. The energy consumption was reduced by 31.73–45.90% in their experiments.

Besides regression models, some studies have also applied classification algorithms to investigate and categorize the impacts of driving behavior on fuel consumption. The targets in classification models are discrete, so the energy efficiency assessment indicators were divided into several categories, e.g., efficient skilled driver, moderate skilled driver, efficient novice driver, and moderate novice driver [14]. Lu et al. [7] applied Gaussian mixture models (GMM) to identify the driving modes of different drivers for strategy optimization, and the planning strategy in this study saved 15.81% of the fuel compared with the aggressive driver. Chen et al. [8] established a bus fuel consumption classification model based on the decision tree. Various driving strategies are classified into three groups, including low-, medium-, and high fuel consumption. Three energy-efficient driving recommendations are provided based on the classification results. Rettore et al. [35] applied K-means clustering to analyze the correlation between a vehicle's speed and the engine's number of revolutions per minute (RPM). Their recommendations of the best gear considering speed and torque enabled an energy saving of up to 21%.

Moreover, combined learning frameworks can take advantage of both the classification and regression models. In these cases, the classification algorithms are used to identify different driving patterns, and the regression models estimate the energy consumption based on the selected algorithms. For instance, the random forest was utilized in [32] to classify multiple segments of the driving process into three patterns, and then bagging-REPTree was used to predict the fuel consumption in each pattern. Furthermore, experiments demonstrated that the proposed framework can reach more than 7% $CO_2$ emissions reduction. For example, Chang et al. [22] identified the typical driving primitives through the hidden Markov model (HMM) and k-means clustering, and obtained the energy efficiency and emission evaluation result through the linear regression model.

3.5.2. Reinforcement Learning for Energy-Efficient Driving

Reinforcement learning (RL) is mainly used to optimize agents' actions in an environment by repeated interactions to maximize the cumulative reward [63,64]. The establishment of an RL environment is typically based on the Markov decision process (MDP) which is a discrete-time stochastic control process [65]. An RL framework is agent-based, with the agent being the targeted vehicle whose fuel consumption is targeted to be minimized, and the surrounding vehicles, roads, signals, traffic rules, and other parameters constitute the environment [36]. The general scenario is visualized in Figure 4. The roadside units represent the facilities that affect the driving styles, including bus stations, schools, etc.

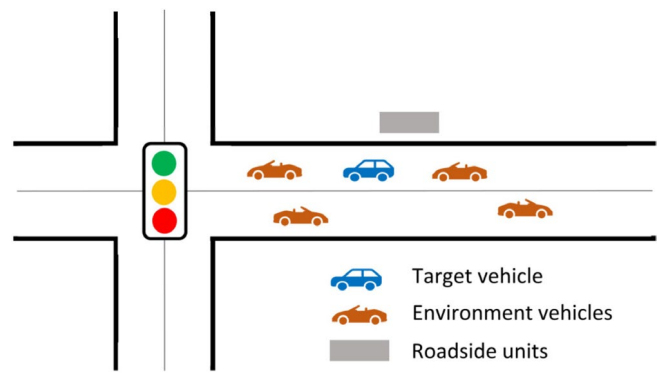

**Figure 4.** RL energy-efficient driving scenario.

Compared with other model-based simulation schemes, the RL-based energy-efficient driving method improves the accuracy and generalization [4]. On one hand, other models can deal with specific driving scenarios like intersection passing and car following with their different analysis strategies. On the other hand, the problems in the current non-RL mode need to be decomposed into sub-blocks for numerical solutions when the scenario is too complex. The RL frameworks overcome such drawbacks by directly learning the optimal value or policy in a black box, regardless of the environment's complexity.

In energy-efficient driving cases, RL has been applied to simulate the driving scenarios and make the optimized suggestions or has been directly embedded into the vehicle control systems [38]. For instance, Guo et al. [12] proposed a hybrid deep Q-learning and policy gradient algorithm, considering both the longitudinal acceleration and the lateral lane-changing operations for vehicles along multi-lane signalized corridors. The longitudinal and lane-changing fuel-saving strategies are learned based on the developed algorithm, which reduces the energy consumption by approximately 46%. Lee et al. [38] developed a model-based RL model to optimize the speed trajectory using the road slope of the driving cycle. The proposed model exhibits the fuel saving performance of 1.2–3.0% compared with existing cruise control and dynamic programming methods. Liu et al. [41] designed a multi-light road environment for training and used the twin delayed deep deterministic policy gradient algorithm (TD3) as a learning mechanism. The strategy based on multi-light trained models saved over 10% energy compared with the optimization results from single-light models. Zhu et al. [47] proposed a Q-learning-based energy-efficient driving optimization approach, and it obtained the global optimum without a complicated control framework.

## 4. Discussion

### 4.1. Pipelines of Energy-Efficient Driving Problem

This in-depth literature review reveals that recent energy-efficient driving research is mainly based on big data analysis and machine learning algorithms. The methodologies can be divided into two groups: (1) supervised/unsupervised learning algorithms are developed to analyze the performance of different energy-efficient driving strategies. Meanwhile, the best-practice energy-efficient driving style and the important driving-behavior factors are selected based on the classification and identification results; (2) the driving scenario is established in the RL environment, and the driving strategy is optimized in the built scenario through deep learning frameworks.

Figure 5 illustrates the pipeline of supervised/unsupervised learning models concluded from the literature. First, the driving data are collected from different sources as shown in Table 4. The data in such research are mostly acquired by CANbus [22], OBD devices [33], and other independent sensors [14] in real road experiments. Afterwards, the significant driving-behavior features are extracted based on the feature selection algorithms (e.g., SHAP [37] and LIME [66]) or their contributions in some white-box prediction models (e.g., linear regression [21] and random forest [10]). Next, the regression models can be applied to predict fuel consumption and select the most efficient driving strategy after comparison, or the classification frameworks are conducted to classify the driving schemes for choosing the best practices in driving styles.

Figure 6 demonstrates the RL-based energy-efficient driving pipeline. The embedded driving scenario should be simplified considering the hardware performance. Hence, the real-world data cannot be directly used to build an RL environment. In the cases [21,29,42], limited aspects in real road and traffic conditions are considered for RL simulation, and some data are collected from the driving simulators. Normally, the traffic signals, road slope, and lanes are set in the RL environment, and the speed and acceleration are optimized to reach the minimum fuel consumption. The driving styles are optimized iteratively by the training module and the best practice is provided when the reward is maximized.

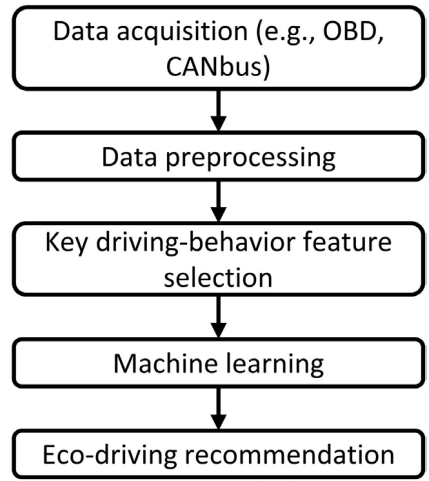

**Figure 5.** Supervised/unsupervised learning-based pipeline.

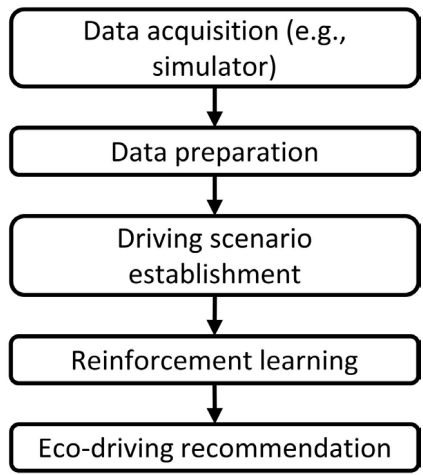

**Figure 6.** RL-based pipeline.

The two mentioned ML-based frameworks contain specific properties. The supervised/unsupervised learning models are capable of processing real-world datasets with a large number of variables, which provide actual and multiple road and traffic information. Furthermore, the important driving-behavior variables concerning energy efficiency are detected in some pipelines, so detailed fuel-saving recommendations can be offered depending on each factor. However, the selected driving strategy may not be the best practice because it is among the limited candidates in the designed experiments.

The quality of the recommendation is highly related to the experiment designers. The RL is an optimization framework, so it is able to provide more convincing energy-efficient driving recommendations compared with the supervised/unsupervised learning algorithms. For instance, the deep Q-learning framework utilizes deep neural networks to approximate values, maximizing the reward [47]. Moreover, the road and vehicle parameters in the simulation environment can be designed based on the research demands, and the irrelevant factors can be removed or set as constant. However, the drawback of RL is that the complicated traffic network is not easy to simulate due to the computational limitation. In the applications, most articles conduct research on scenarios containing a few constraints, e.g., signalized intersections [12] or lane changing [36]. Except for that, the actual random road damage and weather changes are not able to be simulated, so the optimization strategies may not match the real driving conditions well.

*4.2. Data Sources*

The driving behavior data are collected through virtual simulators or naturalistic experiments. The supervised/unsupervised learning models process real-world data, whereas the RL models utilize simulation data in most cases.

Experimental data are more convincing and richer compared with virtual data owing to the complicated actual road and traffic conditions. However, the quality of real-world data are not guaranteed all the time, depending on the data collection devices and experiment execution. Even in the same experiment, each driver's data collector must be the same model to ensure similar systematic errors, otherwise the unpredictable errors could result in outliers and errors that are not easily corrected. In addition, the accuracy of the devices should be validated before the experiments to reduce errors. Furthermore, the experiments always cover several drivers with multiple driving styles, and the samples should be able to reflect the entire population, which is hard to realize. Moreover, the driving styles in the experiments may not involve the most energy efficiency due to the limited participants, so the optimal recommendation of ecological driving may not be reached in the analysis. Therefore, a large number of drivers with driving styles meeting the experimental requirements should participate in the experiments.

Simulated data are more stable than real-world data since they are not influenced by random environmental noise, which increases the success rate and the efficiency of the experiments. Moreover, extreme conditions like sudden braking can be tested in the simulation environment, ensuring driver safety. Multiple and changeable weather conditions can be set flexibly in simulations as well, of which the data cannot be collected easily in real-world experiments in that the weather is uncontrollable. However, the simulation environment is not capable of replicating the actual scenario because the parameters are simplified according to the corresponding assumptions and mathematical models. To overcome this problem, the parameters should be set following the real road conditions and traffic flows. In the meantime, the simulating software that can build realistic scenes needs to be selected for data collection.

*4.3. Recommendation on Energy-Efficient Driving Styles*

Based on the detailed analysis of the reviewed 37 articles, the relationships between driving behaviors and energy consumption, this paper lists nine recommendations for energy-efficient driving behaviors, including driver characteristics, driving behavior factors, and safety rules. Recommendations 1–4 guide the drivers to select and adjust the proper vehicle parameters, and recommendations 5–7 suggest energy-efficient driving styles in different driving scenarios. The subjective suggestions for different types of drivers and employers are provided in recommendations 8–9.

1. Speed: Accelerating action assumes more fuel than driving at a steady speed; however, the speed always fluctuates within a certain range in a long-term driving period. The drivers are recommended to keep the speed fluctuating at the range of ±1 km/h at the beginning of the cruise period for at least 20 s [13]. Moreover, extreme low and high cruise speeds both result in high fuel consumption, so a moderate velocity should be chosen as an appropriate cruise speed.
2. Acceleration: There exists a strong correlation between speed, acceleration, and fuel consumption [67]. Accelerating at high speeds causes more energy consumption than at a low speeds, and rapidly changing the depth of the gas pedal wastes energy as well due to the unmatched engine RPMs. Therefore, drivers should shift to a low gear during the accelerating process, with a moderate throttle and low engine RPMs.
3. Engine: As mentioned above, a higher engine RPM leads to a higher fuel consumption, so the RPM should be controlled at an appropriate value matching the gears. Moreover, the idle time needs to be minimized because the fuel consumed during idling periods is not converted into the mechanical energy of the vehicle's motion. Therefore, drivers cannot keep the engine on when parking the vehicle unless encountering a red signal or other emergencies.

4. Gear selection: In manual transmission cars, the gear should be selected based on the torque and force to maximize the usage of the energy provided by gasoline combustion, which represents that the gear of the transmission box should be changed at a proper speed. The gear shift timing is provided by the vehicle manufacturer, depending on the model of the transmission box and engine.
5. Traffic signals: When the vehicle is about to cross the intersection and observes a switch to a green signal, it either maintains constant cruising speed or low acceleration to avoid the red signal [29]. Aggressively speeding up helps the vehicle escape the red signal but compromises fuel efficiency. While facing the red signal, the driver needs to control the braking pedal to slow down the vehicle to the minimum velocity, and then observe the signal when deciding to stop or cross the intersection slowly.
6. Lane change: Rapid and frequent lane changing consumes much fuel in that it is a velocity-changing process. As recommendation (1) mentioned, the drivers ought to do their utmost to avoid velocity change. When it is necessary to change lanes to turn right/left or overtake other vehicles, the driver should turn the steering wheel smoothly and slowly without aggressive acceleration.
7. Weather: The visibility and road humidity influence the safe speed limit and braking distance [68]. The energy-efficient driving strategy should be adjusted and optimized by considering different weather conditions to promote safe driving. Warning signs can be implemented in the traffic network to assist the drivers in averting energy-efficient but unsafe actions.
8. Types of drivers: Drivers can be classified into the skilled cluster and novice cluster owing to their distinct driving habits and response ability [14]. Skilled drivers save more energy than novice ones due to their lower pedal frequency. However, they can easily become aggressive drivers, stepping on the gas pedal rapidly. They have room to improve their acceleration strategy for lower energy consumption. The novice drivers are mostly overly cautious and use the pedal frequently. Hence, they are recommended to practice more to improve their proficiency, especially to reduce the pedal usage and increase the pedal strength.
9. Work conditions of drivers: This recommendation is specially provided to the drivers' employers. The driver's mood has a significant impact on the driving style, and a negative temper may increase fuel consumption [8]. Good work conditions can mitigate drivers' pressure and improve their mood to avoid frequent pedal usage and reduce high-speed driving. Therefore, employers should improve the drivers' salaries, limit their working hours, and keep them away from drowsy driving.

Furthermore, some publications leverage historical driving data to directly recommend energy-efficient driving styles. In contrast, several articles center around reinforcement learning (RL), assuming the integration of intelligent connected vehicles, thereby enabling direct control over vehicle motion. While these RL-focused studies assume automated control, it is noteworthy that optimal speed control strategies generated from the RL training, traditionally associated with manual driving, can also be harnessed for train drivers and guide them toward achieving energy-saving objectives.

## 5. Conclusions

This paper conducts a thorough literature review on energy-efficient driving styles, and analyzes the driving factors influencing energy consumption and state-of-the-art methodologies. It is evident from the literature that AI models are widely applied in the transportation sector to reduce $CO_2$ emissions.

This paper compares and studies the AI models in the recent literature, and concludes the important driving-behavior factors and the corresponding data sources. The literature shows that the driving data collected from simulators or real-world experiments and real-world data are mainly stored and transmitted by meters, CANbus, OBDs, smartphones, and additional sensors installed in the vehicle. One or multiple data sources are used to guarantee the data volume for accurate big data analysis. Furthermore, this paper compares the driving behavior factors and summarizes 11 features, including speed, acceleration, deceleration, pedal, steering, gear, engine, distance, weather, traffic signal, and road parameters, that cover the variables reflecting and influencing driving behaviors.

This paper finds that the AI models proposed and utilized to model the vehicle's energy consumption based on the collected multi-dimensional data can be divided into two groups, including supervised/unsupervised learning algorithms and RL frameworks. In supervised/unsupervised learning algorithms, the classification models are implemented to select the energy-efficiency driving strategies, and the regression models are conducted to predict fuel consumption. Furthermore, the significant driving factors are detected by the white-box algorithms or the feature selection algorithms. RL models are usually established in simulation scenarios to learn the best energy-efficient driving practice by maximizing the reward of the framework that the driving parameters are optimized iteratively.

This paper recommends nine energy-efficient driving styles based on driving behavior factors, driver characteristics, and safety rules. The nine recommendations include four guidelines for the drivers to select and adjust the proper vehicle parameters, three recommendations for energy-efficient driving styles in different driving scenarios, and two subjective suggestions for different types of drivers and employers.

The energy-efficient driving strategies recommendations in this paper are concluded based on the scoping review study. Therefore, further testing and verification study and a series of experiments in the future work are recommended. Furthermore, in future research, AI models are recommended to be embedded into the real-time control system to adjust the driving parameters in a short time. With the development of autonomous driving, the most energy-efficient driving styles can be applied to control the vehicles directly, as the drivers may not follow the recommendation systems, and they may be removed from the future driving systems. To conclude, energy-efficient driving systems should be implemented based on data from various sources and embedded in state-of-the-art AI algorithms for the significant optimization of $CO_2$ emissions in transportation systems.

**Author Contributions:** Conceptualization, Z.M. (Zhipeng Ma), B.N.J. and Z.M. (Zheng Ma); methodology, Z.M. (Zhipeng Ma) and Z.M. (Zheng Ma); software, Z.M. (Zhipeng Ma) and Z.M. (Zheng Ma); validation, Z.M. (Zhipeng Ma), B.N.J. and Z.M. (Zheng Ma); formal analysis, Z.M. (Zhipeng Ma); investigation, Z.M. (Zhipeng Ma) and Z.M. (Zheng Ma); resources, Z.M. (Zhipeng Ma), B.N.J. and Z.M. (Zheng Ma); data curation, Z.M. (Zhipeng Ma) and Z.M. (Zheng Ma); writing—original draft preparation, Z.M. (Zhipeng Ma); writing—review and editing, Z.M. (Zhipeng Ma), B.N.J. and Z.M. (Zheng Ma); visualization, Z.M. (Zhipeng Ma); supervision, B.N.J. and Z.M. (Zheng Ma); project administration, B.N.J.; funding acquisition, B.N.J. All authors have read and agreed to the published version of the manuscript.

**Funding:** This work was funded by the Energy Technology Development and Demonstration Programme (EUDP) in Denmark under the project EUDP 2021-II Driver Coach [case no. 64021-2034].

**Data Availability Statement:** Data will be made available on request.

**Conflicts of Interest:** The authors declare no conflicts of interest.